\documentclass[a4paper,twoside]{article}

\usepackage{epsfig}
\usepackage{subcaption}
\usepackage{calc}
\usepackage[ruled,vlined,linesnumbered]{algorithm2e}
\usepackage{amssymb}
\usepackage{amstext}
\usepackage{amsmath}
\usepackage{lipsum}
\usepackage{graphicx}
\usepackage{tabularx}
\usepackage{amsthm}
\usepackage{multicol}
\usepackage{pgfplots}
\usepackage{multirow}
\usepackage{pslatex}
\usepackage{tikz}
\usepackage{apalike}
\usepackage{SCITEPRESS}     
\usepackage{xcolor}
\usepackage{dsfont}
\usepackage{float}
\usepackage{bm}
\SetKwInput{KwInput}{Input}
\SetKwInput{KwOutput}{Output}

\theoremstyle{definition}
\newtheorem{definition}{Definition}[section]

\newcommand{\set}[1]{\left\{ \left. #1 \right. \right\}}
\newcommand{\paren}[1]{\left( \left. #1 \right. \right)} 
\newcommand\blfootnote[1]{%
  \begingroup
  \renewcommand\thefootnote{}\footnote{#1}%
  \addtocounter{footnote}{-1}%
  \endgroup
}
\newcolumntype{Y}{>{\centering\arraybackslash}X}

\usetikzlibrary{patterns, positioning}
\usetikzlibrary{shapes.geometric, arrows}
\tikzstyle{io} = [rectangle, rounded corners, minimum width=2cm, minimum height=1cm,text centered, draw=black, fill=red!30]
\tikzstyle{estimate} = [trapezium, trapezium left angle=70, trapezium right angle=110, minimum width=2cm, minimum height=1cm, text centered, draw=black, fill=blue!30]
\tikzstyle{process} = [rectangle, minimum width=2cm, minimum height=1cm, text centered, draw=black, fill=orange!30]
\tikzstyle{decision} = [diamond, minimum width=2cm, minimum height=1cm, text centered, draw=black, fill=green!30]
\tikzstyle{arrow} = [thick,->,>=stealth]

\begin{document}

\title{Expert-Guided Symmetry Detection in Markov Decision Processes}

\author{\authorname{Giorgio Angelotti,\sup{1,2} Nicolas Drougard\sup{1,2} and Caroline P. C. Chanel\sup{1,2}}
\affiliation{\sup{1}ISAE-SUPAERO, University of Toulouse, France}
\affiliation{\sup{2}ANITI, University of Toulouse, France}
\email{\{name.surname\}@isae-supaero.fr}
}

\abstract{Learning a Markov Decision Process (MDP) from a fixed batch of trajectories is a non-trivial task whose outcome's quality depends on both the amount and the diversity of the sampled regions of the state-action space. Yet, many MDPs are endowed with invariant reward and transition functions with respect to some transformations of the current state and action. Being able to detect and exploit these structures could benefit not only the learning of the MDP but also the computation of its subsequent optimal control policy. In this work we propose a paradigm, based on Density Estimation methods, that aims to detect the presence of some already supposed transformations of the state-action space for which the MDP dynamics is invariant. We tested the proposed approach in a discrete toroidal grid environment and in two notorious environments of OpenAI's Gym Learning Suite. The results demonstrate that the model distributional shift is reduced when the dataset is augmented with the data obtained by using the detected symmetries, allowing for a more thorough and data-efficient learning of the transition functions.}

\onecolumn \maketitle \normalsize \setcounter{footnote}{0} \vfill
\section{\uppercase{Introduction}}
\label{sec:introduction}
\vspace{-2mm}

\blfootnote{Preprint - Accepted to ICAART 2022}
Model-based Offline Reinforcement Learning (ORL) is the branch of Machine Learning that first fits 
a dynamical model and then obtains a behavioural (optimal) system control policy with the aim of maximizing a predetermined utility function \cite{SUTTON1990216}. The model learning phase is usually based on  
a finite batch of 
trajectories
 followed by the system. 
Although it is always possible to compute confidence intervals on the parameters of the learnt categorical distribution, it is hard to determine the amount of data needed for the model optimisation to return a sufficiently improved policy.

Following \cite{Levine2020OfflineRL} the statistical distance between reality and the learnt model is defined as \textit{distributional shift}. Thus, in order to be able to compute reliable strategies that can be applied to real world scenarios (e.g. planning in healthcare, autonomous driving) the distributional shift must be minimal. In consequence, limiting it by considering strategies close to the one used for data collection, is the main challenge of ORL \cite{Levine2020OfflineRL}. Indeed,  
such strategies are more likely to drive the system in areas of the state-action space whose transitions have been 
thoroughly explored in the batch, allowing a better estimation of the model than in other areas.

Interestingly, the detection of symmetries in physics has always provided greater representability and generalization power to the learnt models \cite{Gross14256}. Moreover, 
symmetry detection can potentially provide knowledge about what will be the time evolution of a system state that has never been explored before. Such an idea can be explored to allow a more data efficient model learning phase for ORL taking advantage of the knowledge of regularities 
in the system dynamics that repeat themselves in different situations, 
independently of the initial conditions.

In this context, a method to detect symmetries for ORL 
relying on expert guidance is presented.
The described method is built in the classical framework of discrete time Markov Decision Processes (MDPs) 
for both discrete and continuous state-action spaces \cite{kolobov2012planning}.

\vspace{-2mm}
\subsection{Illustrative example}
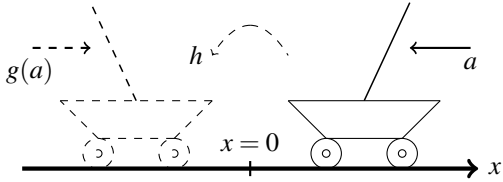
\begin{figure}
\centering
    \begin{tikzpicture}
\draw[->,ultra thick] (-3,0)--(3,0) node[right]{$x$};
\draw [thick] (0, -0.1) -- (0, 0.1) node[above]{$x=0$};
\draw (1, 0.2) circle (0.2cm);
\draw (2, 0.2) circle (0.2cm);

\draw (1, 0.2) circle (0.05cm);
\draw (2, 0.2) circle (0.05cm);

\draw (1, 0.4) -- (0.5, 0.9);
\draw (2, 0.4) -- (2.5, 0.9);
\draw (1, 0.4) -- (2, 0.4);
\draw (0.5, 0.9) -- (2.5, 0.9);

\draw [line width=0.02 cm] (1.5, 0.9) -- (2.1, 2.2);

\draw [<-, thick] (2.1, 1.6) -- (2.9, 1.6) node[below]{$a$};

\draw [dashed] (0, 1.9) parabola (0.5, 1.5);
\draw [->, dashed] (0, 1.9) parabola (-0.5, 1.5) node[above, left]{$h$};

\draw [dashed] (-1, 0.2) circle (0.2cm);
\draw [dashed] (-2, 0.2) circle (0.2cm);

\draw [dashed] (-1, 0.2) circle (0.05cm);
\draw [dashed] (-2, 0.2) circle (0.05cm);

\draw [dashed] (-1, 0.4) -- (-0.5, 0.9);
\draw [dashed] (-2, 0.4) -- (-2.5, 0.9);
\draw [dashed] (-1, 0.4) -- (-2, 0.4);
\draw [dashed] (-0.5, 0.9) -- (-2.5, 0.9);
\draw [line width=0.02 cm, dashed] (-1.5, 0.9) -- (-2.1, 2.2);

\draw [<-, thick, dashed] (-2.1, 1.6) -- (-2.9, 1.6) node[below]{$g(a)$};
    \end{tikzpicture}
    \caption{The cart in the right is a representation of a CartPole's state $s_t$ with $x_t > 0$ and action $a_t=\leftarrow$. The dashed cart in the left is the image of $(s_t,a_t)$ under the transformation $h$ which inverses state $f(s)=-s$ and action $g(a)=-a$.}\label{fig:cartpole}
\end{figure}

\vspace{-4mm}
One of the most emblematic examples is checking whether or not a dynamical system is symmetric with respect to some particular transformation of the system of reference. 
For instance, consider the well-known CartPole Reinforcement Learning (RL) domain of the OpenAI's Gym Learning Suite \cite{brockman2016openai}. 
In CartPole the purpose of the automated agent is to avoid a rotating pole situated on a sliding cart to fall down due to the gravitational acceleration. The state of the system is expressed as a tuple $s=(x, v, \alpha, \omega)$ where $x$ is the position of the cart with respect to a horizontal track upon which it can slide, $v$ is its longitudinal velocity, $\alpha$ is the angle between the rotating pole and the axis pointing along the direction of the gravitational acceleration, and $\omega$ the angular velocity of the pole. The agent can push the cart left ($\leftarrow$) or right ($\rightarrow$) at every time step (in the negative or positive direction of the track) providing to the system a fixed momentum $|p|$. A pictorial representation of a state-action pair $(s_t,a_t)$ can be found in Figure \ref{fig:cartpole}.

Let us suppose there exists a function $h: S\times A \rightarrow S\times A$ that maps a state-action pair $(s_t, a_t)$ to $\big(f(s_t), g(a_t)\big)$, where $f: S \rightarrow S$ and $g: A \rightarrow A$, such that the dynamics of the pair $(s,a)$ is the same as the one of $h(s,a)$. Note that in this example, the system dynamics is symmetric with respect to a flip around the vertical axis. In other words, its dynamics is invariant by multiplication by minus one, assuming that if $a = \leftarrow$ then $g(a) = -a = \rightarrow$ and vice versa.
Indeed, if the state-action pair $(s_t,a_t)$ leads to the state $s_{t+1}$, this property will imply that $h(s_t,a_t)=(-s_t,-a_t)$ leads to $f(s_{t+1})=-s_{t+1}$. 

When learning the dynamics from a finite batch of experiences (or trajectories),
resulting in a set of transitions $\mathcal{D} = \big\{(s_i, a_i, s_i') \big\}_{i=1}^n$ with $n \in \mathbb{N}$ the size of the batch, we might for instance fit a function $\hat{s}(s, a) = s'$ with the aim of minimizing a loss, e.g. the Mean Squared Error. However, imagine that in the batch $\mathcal{D}$ there were many transitions regarding the part of the state-action space with $x>0$ and very few with $x<0$. Unfortunately, we may learn a good model to forecast what will happen when the cart is at the right of the origin and a very poor model at its left side. We can then suppose that also the optimal control policy will perform well when $x>0$ and poorly when $x<0$.

Nevertheless, if it were possible to be confident of the existence of the symmetry $f(s) = -s$ and $g(a) = -a$ 
(where the opposite of the action $a$ is the transformation stated above), we might extend the batch of experiences without additional interaction with the system, and then improve the accuracy of the model also to the regions where $x<0$. 

\subsection{Definition of the problem and related work}
\vspace{-2mm}
How can we automatically detect these sorts of symmetries in the batch of experiences?

In this work we propose a paradigm to check whether a pre-alleged 
structure of this kind is present in the batch. We build our approach upon the Markov Decision Process (MDP) framework \cite{kolobov2012planning}, whether discrete or continuous state and action spaces can be considered, although hypothesizing a non stochastic dynamics.


The detection of structures and abstractions in MDPs has been widely studied in the literature. In \cite{dean1997model,givan2003equivalence} the notions of MDP homomorphism (structure-preserving maps between the original MDP and one characterized by a factored representation) and stochastic bisimulation (are introduced and used to automatically partition the state space of an MDP and to find aggregated and factored representations. In \cite{ravindran2001symmetries} the previous works 
on state abstractions have been extended to include the concept of symmetry, and in \cite{ravindran2004approximate} approximate homomorphisms are considered. 

Later on, \cite{narayanamurthy2008hardness} showed that the fully automatic discovery of symmetries in a discrete MDP is as hard as verifying whether two graphs are isomorphic. Concurrently, \cite{NIPS2008_6602294b} relaxed the notion of bisimulation to allow for the attainment of performance bounds for approximate MDP homomorphisms. Approximate homomorphisms are of particular interest in continuous state MDPs where a hard mapping to an aggregated representation could be impractical. In this context, \cite{ferns2004metrics} developed a bisimulation pseudometric to extend the concept of bisimulation relation. The automatic discovery of representations using the bisimulation pseudometric has been investigated in recent years using Deep Neural Networks and obtaining theoretical guarantees for such a methodology \cite{ruan2015representation,castro2020scalable,abel2020value}. 

From a more theoretical perspective \cite{li2006towards} classified different kinds of possible state abstractions, showing for each one of them which function would be preserved under the new refinement: the value function, the Q-value function, the optimal policy, and so on. From a different perspective \cite{mandel2016efficient} developed an algorithm that aims to cluster MDPs states in a Bayesian sense in order to solve the MDP in a more data efficient way, even when an underlying homomorphic or symmetric structure is not present.

Recently, \cite{10.5555/3398761.3398926} used a contrastive loss function that enforces
action equivariance on a to be learnt representation of an MDP. Their approach resulted in the automatic learning of a structured latent space which can be used to plan in a more data efficient fashion. Finally, \cite{NEURIPS2020_2be5f9c2} introduced MDP Invariant Networks, a specific class of Deep Neural Network architectures that guarantees by construction that the optimal MDP control policy obtained through other Deep RL approaches will be invariant under some set of symmetric transformations and hence providing more sample efficiency to the baseline when the symmetry is actually present.

\subsection{Contribution}
\vspace{-2mm}
In summary, the literature can be divided in two categories: 1) learning a representation using only or the data contained in the batch or some a priori knowledge about the environment; 2) exploiting an alleged symmetry to obtain a behavioural policy that converges faster to the optimal one.

The present work gets inspiration from all the reported literature, but rather than automatically looking for a state space abstraction or directly exploiting a symmetry to act in the world it focuses on checking whether the dynamical model to be learnt is symmetric with respect to some transformation that is presumed by other means. In other words, such transformation can be given by external expert knowledge or proposed by the user insight and its existence is verified by estimating the chance of its occurrence with machine learning tools.

In order to verify the presence of a symmetry, a preliminary estimate of the transition model is needed.
More specifically, we first perform a probability mass function (pmf) estimation or a probability density function (pdf) estimation of the transitions in the batch $\mathcal{D}$ depending on the typology of the MDP we are tackling (respectively discrete or continuous).
In the discrete case this amounts to learning a set of categorical distributions. 
In the continuous case we opt for using Normalizing Flows \cite{DBLP:journals/corr/DinhKB14,kobyzev2020normalizing}, a deep neural network architecture that allows to adapt and estimate a pdf while being always able to exactly compute the density value for new samples. In this way, once a pmf/pdf has been estimated from the batch of transitions $\mathcal{D}$ we can compute the probability of an alleged symmetric transition that is supposed to be sampled from the same distribution.

When the probability (or the density in the continuous case) is bigger than a given threshold for a high fraction of samples, we decide to \textit{trust} in the presence of this alleged symmetry and hence augment the batch by including the symmetric transitions.

In the end the dynamics of the model is learnt over the augmented data set. When the approach detects a symmetry that is really present in the true environment then the accuracy of the learnt model increases, otherwise the procedure could also result in detrimental effects.


The paper is organized as follows: in Section \ref{sec:background} we provide the definitions of MDP, MDP homomorphism, symmetric transformation and pmf/pdf estimation; next, in Section \ref{sec:approach} the two algorithms for expert-guided detection of symmetries are presented along with a discussion of the limitations of both (these are the main contributions); results in proof-of-concept environments are shown in Section \ref{sec:experiments}; conclusion and future perspective are illustrated in Section \ref{sec:conclusion}.
\vspace{-6mm}
\section{\uppercase{Background}}
\vspace{-2mm}
\label{sec:background}
In this section, we introduce the basic concepts about MDPs, homomorphisms and symmetries.

\begin{definition}[Markov Decision Process]
A discrete-time MDP \cite{bellman1966dynamic} is a tuple $\mathcal{M} = (S, A, R, T, \gamma)$, where $S$ is the set of states, $A$ is the set of actions, $R : S \times A \rightarrow \mathbb{R}$ is the
reward function, $T : S \times A \rightarrow Dist(S)$ is the transition pdf, and $\gamma \in [0, 1)$ is the discount factor. At each time step, the agent observes a state $s=s_t \in S$, takes an action $a=a_t \in A$ drawn from a policy $\pi : S \times A \rightarrow [0, 1]$, and with probability $T(s'|s, a)$ transits to a
next state $s'=s_{t+1}$, earning a reward $R(s_t, a_t)$. The value function of the policy $\pi$ is defined as: $V_{\pi}(s) = \mathbb{E}_{\pi}\big[\sum_{t=0}^{\infty}\gamma^{t} R(s_t, a_t)|s_0 = s\big]$. The optimal value function $V^{*}$ is the maximum of the previous expression over every possible policy $\pi$. When $\forall (s,a) \in S\times A$, $T$ dictates the transition to one and only state $s'\in S$ the MDP is said to be deterministic or non stochastic.
\end{definition}

A homomorphism from a dynamic system $\mathcal{M}$ to a dynamic system $\mathcal{M'}$ is a mapping that preserves $\mathcal{M}$’s dynamics, while in general eliminating some of the details of the full system $\mathcal{M}$.

\begin{definition}[MDP Homomorphism]
An MDP homomorphism $h$ \cite{ravindran2004approximate} from an MDP $\mathcal{M} = (S, A, T, R, \gamma)$ to an MDP $\mathcal{M'} = (S', A', T', R', \gamma)$ is a surjection from $S\times A$ to $S'\times A'$
, defined by a tuple of surjections $(f, g)$, with $h(s, a) = \big(f(s), g(a)\big)$, where $f : S \rightarrow S'$ and $g: A \rightarrow A'$ such that $\forall (s,s')\in S^2, a \in A$:
\begin{equation}
    T'\big(f(s), g(a), f(s')\big) = \sum_{s'' \in [s']_f} T(s,a,s''),
\end{equation}
\begin{equation}
    R'\big(f(s), g(a)\big) = R(s,a).
\end{equation}
where $[s']_f = f^{-1} \Big( \big\{ f(s') \big\} \Big)$, \textit{i.e.} $[s']_f$ is the set of states for which the application of $f$ results in the state $f(s') \in S'$.
\end{definition}
\begin{definition}[MDP Symmetry]
Let $k$ be a a surjection from an MDP $\mathcal{M}$ to itself defined by the tuple of surjections $(f, g, l)$ where $f : S \rightarrow S$, $g : A \rightarrow A $ and $l: S \rightarrow S$. The transformation $k$ is a symmetry if $\forall (s,s')\in S^2$, $a \in A$ both the dynamics $T$ and the reward $R$ are invariant with respect to the transformation induced by $k$:
\begin{equation}
\label{eq:dyn}
T\big(f(s),g(a),l(s')\big) = T(s,a,s'),
\end{equation}
\begin{equation}
R\big(f(s),g(a)\big) = R(s,a).
\end{equation}
\end{definition}
If $\mathcal{M'} = \mathcal{M}$ and $l = f$ then an MDP symmetry is also an MDP homomorphism.

In this work we will focus only on symmetries with respect to the dynamics and not the reward.
In particular, we will check for transformations $k$ that map a whole transition $k(s,a,s') \rightarrow \big(f(s),g(a),l(s')\big)$.
Therefore the only condition that will be examined and fulfilled is Equation \ref{eq:dyn}.
Detecting the said structure will help reduce the distributional shift between the true environment dynamics and the one learnt from experience.

\paragraph{Probability Mass Function Estimation for Discrete MDPs.}
Let $\mathcal{D} = \{(s_i, a_i, s_i')\}_{i=1}^n$ be a batch of recorded transitions. Performing mass estimation over $\mathcal{D}$ amounts to compute the probabilities that define the unknown categorical transition distribution $T$ by estimating the frequencies of transition in $\mathcal{D}$.
In other words we compute: 
\begin{equation}
    \hat{T}(s,a,s') = \begin{cases}
    \dfrac{n_{s,a,s'}}{\sum_{s'} n_{s,a,s'}} &\text{if $\sum_{s'} n_{s,a,s'} > 0$,}\\
|S|^{-1} &\text{otherwise;}\end{cases}
\end{equation}
where $n_{s,a,s'}$ is the number of times the transition $(s_t = s, a_t = a, s_{t+1} = s') \in \mathcal{D}$.

\paragraph{Probability Density Function Estimation for Continuous MDPs.}
Performing density estimation over $\mathcal{D}$ amounts to finding an analytical expression for the probability density of a transition $(s, a, s')$ given $\mathcal{D}$: $\mathcal{L}(s, a, s' | \mathcal{D})$. Normalizing flows \cite{DBLP:journals/corr/DinhKB14,kobyzev2020normalizing} allow defining a parametric flow of continuous transformations that reshapes a known pdf (e.g. a multivariate gaussian) to one that best fits the data. Since the transformations are known, the Jacobians are computable at every step and the probability value can always be assessed. \cite{DBLP:journals/corr/DinhKB14}.
\section{\uppercase{Expert-Guided Detection of Symmetries}}
\vspace{-3mm}
\label{sec:approach}
We propose a paradigm to check whether the dynamics of a to be learnt (i) deterministic  discrete MDP (see Algorithm \ref{algo:disc}) or (ii) a deterministic continuous MDP (see Algorithm \ref{algo:cont}) is endowed with the invariance of the dynamics with respect to some transformation.

Let $\mathcal{M}$ be a deterministic MDP, let $\mathcal{D}$ be a batch of pre collected transitions and let $k$ be an alleged symmetric transformation of $\mathcal{M}$'s dynamics.

\begin{algorithm}
\SetAlgoLined
\KwInput{Batch of transitions $\mathcal{D}$, $\nu \in (0,1)$ percentage threshold, $k$ alleged symmetry}
\KwOutput{Possibly augmented batch $\mathcal{D}\cup \mathcal{D}_k$}
 $\hat{T} \leftarrow$ Most Likely Categorical pmf from ($\mathcal{D}$)\\
 $\mathcal{D}_k = k(\mathcal{D})$\\
$\nu_{k} = \displaystyle \dfrac{1}{|\mathcal{D}_k|}\sum_{(s,a,s') \in \mathcal{D}_k} \mathds{1}_{\{\hat{T}(s'|s,a)=1\}}$\\
 \eIf{$\nu_k > \nu$}{
  \textbf{return}
  $\mathcal{D}\cup \mathcal{D}_k$}
   {\textbf{return} $\mathcal{D}$
  }

 \caption{Symmetry detection and data augmenting in a deterministic discrete MDP}
 \label{algo:disc}
\end{algorithm}

\begin{figure}[b]
\centering
\includegraphics[width=\columnwidth]{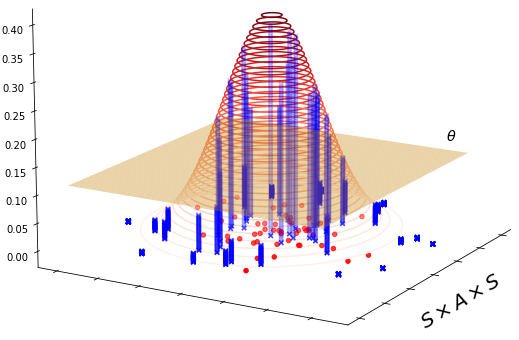}
\caption{Intuition for the continuous case. The $xy$ plane is the space of transitions $S \times A \times S$, the $z$ axis is $\mathcal{L}$, the value of the probability density of a given transition. The red points represent $\mathcal{D}$, the blue crosses $\mathcal{D}_k$ for a given transformation $k$. We display as a red contour plot the pdf $\mathcal{L}$ learnt in Line 1 of Algorithm \ref{algo:cont}. The orange hyperplane has height $\theta$ which is the threshold computed in Line 3 of Algorithm \ref{algo:cont}. The blue vertical bars have as height the value of $\mathcal{L}$ evaluated for that specific transition. The algorithm counts the fraction $\nu_k$ of samples (blue crosses) which have a vertical bar higher than the hyperplane.}\label{fig:3dfig}
\end{figure}
\begin{figure*}[t]
\centering
    
    \begin{tikzpicture}[node distance=4cm]
    
    \node (start) [io,text width=2cm] {Estimate transitions};
    \node (estimate1) [estimate, right of=start,text width=2cm] {Transform data with $k$};
    \node (pro1) [process, right of=estimate1,text width=3cm] {Evaluate probabilities};
    \node (aug) [io, right of=pro1,text width=2cm] {Augment or not $\mathcal{D}$};
    \draw [arrow] (start) -- (estimate1);
    \draw [arrow] (estimate1) -- (pro1);
    \draw [arrow] (pro1) -- node[anchor=south] {$\nu_k > \nu$?} (aug);
    \end{tikzpicture}
        \caption{Pseudo flow chart of Algorithms \ref{algo:disc} and \ref{algo:cont}.}\label{fig:flow}
\end{figure*}
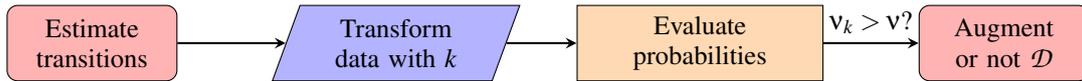

In order to check whether $k$ can be considered or not as a symmetry of the dynamics, in the discrete case we will first estimate the most likely set of transition categorical distributions $\hat{T}$ given the batch of transitions (Line 1, Algorithm \ref{algo:disc}) while in the continuous case we will estimate the density of transition in the batch obtaining the probability density $\mathcal{L}(s,a,s'|\mathcal{D})$ (Line 1, Algorithm \ref{algo:cont}). In the continuous case we will then compute the density value of every transition $(s,a,s') \in \mathcal{D}$, resulting in a set of real values from $\mathcal{L}$ denoted by $\Lambda$ (Line 2, Algorithm \ref{algo:cont}).
We will select the $q$-order quantile of $\Lambda$ to be a threshold $\theta \in \mathbb{R}$ (Line 3, Algorithm \ref{algo:cont}) that will determine whether we can trust the symmetry to be present, and hence to augment or not the starting batch. Next, we will map any $(s, a, s') \in \mathcal{D}$ to its alleged symmetric image $\big(f(s), g(a), l(s')\big)$ (Line 4, Algorithm \ref{algo:cont}). The map of $\mathcal{D}$ under the transformation $k$ will be denoted as $\mathcal{D}_k$. Let then, $\forall (s,a,s') \in \mathcal{D}_k$, $\mathcal{L}(s,a,s' | \mathcal{D})$ be the probability density of the symmetric image of a transition in the batch. We will assume that the system dynamics is invariant under the transformation $k$ if $\nu_k$, the percentage of transitions whose density is greater than $\theta$, is bigger than the percentage threshold $\nu$ (Lines 5-10, Algorithm \ref{algo:cont}). In the discrete case we will just count how many images of the symmetric transformation $\mathcal{D}_k$ are present in the batch (Line 3, Algorithm \ref{algo:disc}) and then decide whether to perform data augmenting according to both $\nu_k$ and the threshold $\nu$ (Lines 4-8, Algorithm \ref{algo:disc}). If data augmenting is performed, the boosted batch $\mathcal{D} \cup \mathcal{D}_k$ will be returned as output. Figure \ref{fig:3dfig} provides an intuition about Algorithm \ref{algo:cont}. A pseudo representation of the flow chart of both algorithms is shown in Figure \ref{fig:flow}.

\vspace{1cm}
\begin{algorithm}
\SetAlgoLined
\KwInput{Batch of transitions $\mathcal{D}$, $q \in [0,1)$ order of the quantile, $\nu \in (0,1)$ percentage threshold, $k$ alleged symmetry}
\KwOutput{Possibly augmented batch $\mathcal{D}\cup \mathcal{D}_k$}
 $\mathcal{L} \leftarrow$ Density Estimate ($\mathcal{D}$)\\
 $\Lambda \leftarrow$ Distribution $\mathcal{L}(\mathcal{D})$ ($\mathcal{L}$ evaluated over $\mathcal{D}$)\\
 $\theta = q$-order quantile of $\Lambda$\\
 $\mathcal{D}_k = k(\mathcal{D})$\\
 $\nu_{k} = \displaystyle \dfrac{1}{|\mathcal{D}_k|}\sum_{(s,a,s') \in \mathcal{D}_k} \mathds{1}_{\{\mathcal{L}(s,a,s'|\mathcal{D})>\theta\}}$\\
 \eIf{$\nu_k > \nu$}{
  \textbf{return}
  $\mathcal{D}\cup \mathcal{D}_k$}
   {\textbf{return} $\mathcal{D}$
  }

 \caption{Symmetry detection and data augmenting in a deterministic continuous MDP}
 \label{algo:cont}
\end{algorithm}

\vspace{-8mm}
\subsection{Intuition for the discrete case}
\vspace{-2mm}
In the discrete case we are performing a sort of ``proof by induction``. If the majority of symmetric transitions (with respect to the threshold $\nu$) were already sampled, then we can imagine that the other ones were not sampled just because our data set is not big enough. Therefore we trust that the dynamics is symmetric with respect to $k$ and augment the batch.

\subsection{Intuition for the continuous case}
\begin{figure}[H]
\centering
    \begin{tikzpicture}
    \draw [thick] (3,0) ellipse (2cm and 1cm) node[below, right] {$\mathcal{D}$};
    \draw [dashed, rotate=-45, thick] (0,2) ellipse (2cm and 1cm) node {$\mathcal{D}_{k_1}$};
    \draw [dashed] (4,2.1) circle (1cm and 1cm) node {$\mathcal{D}_{k_2}$};
    \begin{scope}
        \clip (3,0) ellipse (2cm and 1cm);
        \fill[pattern=north east lines, rotate=-45] (0,2) ellipse (2cm and 1cm);
    \end{scope}
    \node at (0, -0.5) {$S \times A \times S$};
    \end{tikzpicture}
    \caption{Representation of the support in $S^2\times A$ of the of $\mathcal{D}$, $\mathcal{D}_{k_1}$ and $\mathcal{D}_{k_2}$. $k_1$ and $k_2$ are two different alleged transformations. The shape and the position of the sets is decided according to the value of the log likelihood of the density estimate $\mathcal{L}$ and quantile threshold $\theta$. $\mathcal{D}_{k_1} \cap \mathcal{D} \neq \varnothing$ and $\mathcal{D}_{k_2} \cap \mathcal{D} = \varnothing$. Depending on the user chosen percentage threshold $\nu$, $k_1$ could be a symmetry while $k_2$ no. If $k_1$ is detected as a symmetry then data augmenting $\mathcal{D}$ with $\mathcal{D}_{k_1}$ means training the model over $\mathcal{D} \cup \mathcal{D}_{k_1}$. Notice that the data contained in $\mathcal{D}_{k_1}\backslash\mathcal{D}$ are not present in the original batch $\mathcal{D}$.}\label{fig:sets}
\end{figure}
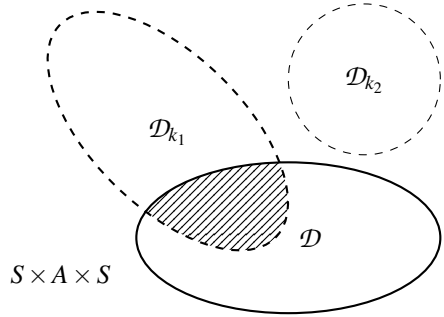

\vspace{-2mm}
The intuition behind the approach in the continuous case is that if in the original batch $\mathcal{D}$ the density of transitions that are ``not so different" from some of the symmetric images $\mathcal{D}_k$ of $\mathcal{D}$, then $\mathcal{L}(\mathcal{D}_k | \mathcal{D})$ will not be ``too small". How small is small when we are considering real valued, continuous pdf? In order to insert a comparable scale we take the threshold $\theta$ to be a $q$-quantile of the set of the estimated density values of the transitions in the original batch $\mathcal{D}$, \textit{i.e.} $\big\{ \mathcal{L}(s,a,s' \mid \mathcal{D}) \mid (s,a,s') \in \mathcal{D} \big\}$.
It goes without saying that since the purpose of Algorithm \ref{algo:cont} is to perform data augmentation, it is necessary to select a small $q$-order quantile, otherwise the procedure would bear no meaning: it would be pointless to augment the batch with transitions that are already very likely in the original one (see Figure \ref{fig:sets}). In this case we won't insert any new information.

\subsection{Limitations}
\vspace{-2mm}
Both in the discrete and in the continuous case the approach is limited by the size and the variety of the data set. It is hard to tell how good an initial batch is, but this aspect could non negligibly affect the output of the paradigm.
The approach in the continuous case is also limited by the impact that the transformation has on the representation of the transition. More in detail, if the distance in Euclidean space between $(s,a,s')$ and $k(s,a,s')$ is small, or if only a small fraction of the total number of components that represent the transition is affected by the transformation then the probability of the image of the alleged symmetry will be high, resulting in a false positive detection. With this in mind, appropriately preprocessing the data before density estimation could enormously affect the outcome of the procedure.


\vspace{-4mm}
\section{\uppercase{Experiments}}
\vspace{-2mm}
\label{sec:experiments}
We test the algorithm in one discrete grid environment with periodic boundary conditions and in two famous environments of OpenAI's Gym Learning Suite: CartPole and Acrobot.

\vspace{-2mm}
\subsection{Setup}
\vspace{-2mm}
We first collect a batch of transitions $\mathcal{D}$ by acting in the environment with a uniform random policy.
We suppose the presence of a symmetry $k$ and we try to detect it using Algorithm \ref{algo:disc} or Algorithm \ref{algo:cont}. 
We report $\overline{\nu}_k$, the average value plus or minus the standard deviation of the quantity $\nu_k$ computed over an ensemble of $N$ different iterations of the procedure (with $N$ different batches $\mathcal{D}$). We do not consider any threshold $\nu$, thus limiting the analysis at the phase of the calculation of $\nu_k$ since the detection or not of a symmetry is highly dependent of an expert-wise domain dependent choice of $\nu$.
In the end we show that detecting the right symmetry really results in learning more accurate models by computing $\overline{\Delta}$, the average of an estimate of the distributional shift $\Delta$ over the various simulations. Since $\Delta$ is measured differently depending on the typology of the environment (discrete or continuous) more details about the performances are illustrated in the next subsections and the results are reported in Table \ref{tab:results}. In Tables \ref{tab:grid-transf}, \ref{tab:cartpole-transf}, \ref{tab:acrobot-transf} we adopted the following contracted notation: let $a_1, a_2, a_3, a_4$ be some actions $\in A$, for space's sake we will use the notations $g(a_1, a_2, \dots) = (a_3, a_4, \dots)$ to indicate $g(a_1) = a_3, g(a_2) = a_4, \dots$, etc.

\subsection{Grid (Discrete)}
\vspace{-2mm}
In this environment the agent can move along fixed directions over a torus by acting with any $ a\in A = \set{\uparrow, \downarrow, \leftarrow, \rightarrow}$.
\begin{figure}[bp!]
\centering
\begin{tikzpicture}[scale=0.8, transform shape]
    \begin{axis}[hide axis]
       \addplot3[surf,
       colormap/blackwhite,
       samples=20,
       domain=0:2*pi,y domain=0:2*pi,
       z buffer=sort]
       ({(2+cos(deg(x)))*cos(deg(y+pi/2))}, 
        {(2+cos(deg(x)))*sin(deg(y+pi/2))}, 
        {sin(deg(x))});
        \addplot3[color=red, thick, ->] coordinates {(-0.85,-0.85,0) (-0.85,-0.85,0.25)} ;
        \addplot3[mark=*,red,point meta=explicit symbolic,nodes near coords] 
coordinates {(-0.85,-0.85,0)[]};
    \end{axis}
\end{tikzpicture}
\caption{Representation of the Grid Environment. The red dot is the position of a state $s$ on the torus. The displacement obtained by acting with action $a=\uparrow$ is shown as a red arrow.}\label{fig:torus}
\end{figure}
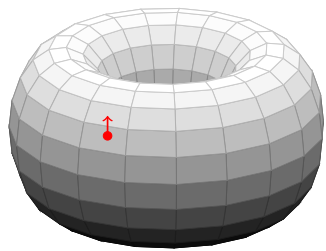
The positions on the torus are the states $s = (i, j)$ and the set $S$ is represented as a grid of fixed size $l$ and periodic boundary conditions (see Figure \ref{fig:torus}). Since there are no obstacles and the dynamics is deterministic this environment is endowed of many symmetric transformations and therefore can serve as an useful proof-of-concept.\\
We tested Algorithm \ref{algo:disc} with six different alleged transformations $k$ in a Grid with size $l = 100$ over $N = 50$ different simulations.
\begin{enumerate}
    \item \textbf{Time reversal symmetry with action inversion (TRSAI).} Assuming that $\downarrow$ is the reverse of $\uparrow$ and $\leftarrow$ is the reverse of $\rightarrow$ we proposed the following transformation: $k = \big(f(s)=s'$, $g(\uparrow, \downarrow, \leftarrow, \rightarrow) = (\downarrow, \uparrow, \rightarrow, \leftarrow)$, $l(s') = s\big)$.
    The symmetric transformation is present in the batch with a percentage $\overline{\nu}_k = 0.6 \pm 0.1$.
    \item \textbf{Same dynamics with action inversion (SDAI).}
    When $k = \big(f(s)=s$, $g \paren{\uparrow, \downarrow, \leftarrow, \rightarrow} = \paren{\downarrow, \uparrow, \rightarrow, \leftarrow}$, $l(s')=s'\big)$ the results clearly show that this transformation is not a symmetry. Indeed in this case $\overline{\nu}_k = 0.0 \pm 0.0$.
    \item \textbf{Opposite dynamics and action inversion (ODAI):} $k = \big(f(s)= s$, $g \paren{\uparrow, \downarrow, \leftarrow, \rightarrow} = \paren{\downarrow, \uparrow, \rightarrow, \leftarrow}$, $l(s') =s' \mp (2, 0) \lor (0, 2)\big)$. In other words we revert the action but also the final state is changed in order to reproduce the correct destination. $\overline{\nu}_k = 0.4 \pm 0.1$ showing that the latter could be a symmetry of the environment.
    \item \textbf{Opposite dynamics but wrong action (ODWA).}
    The alleged transformation is like the one of Point 3, but the action is switched on the wrong axis (e.g. $g(\uparrow) = \rightarrow$). Here $\overline{\nu}_k = 0.0 \pm 0.0$ and not a symmetry.
    \item \textbf{Translation invariance (TI).} $k = \big(f(s)=s', g \paren{\uparrow, \downarrow, \leftarrow, \rightarrow} = \paren{\uparrow, \downarrow, \leftarrow, \rightarrow}, l(s') = s' \pm (1, 0) \lor (0, 1) \big)$. The proposed transformation proposes a displacement with similar effects when the same action is applied to the next state. $\overline{\nu}_k = 0.5 \pm 0.1$. $k$ could be a symmetry.
    \item \textbf{Translation invariance with opposite dynamics (TIOD).} In this case the action is the same as Point 5 but the agent returns to the previous state. With $\overline{\nu}_k = 0.0 \pm 0.0$ the latter is not a symmetry.
\end{enumerate}
\normalsize
\begin{table}[tbh]
\caption{Grid. Proposed transformations and label}\label{tab:grid-transf} \centering
\begin{tabularx}{\columnwidth}{cY}
  \hline
  \noalign{\vskip 0.1mm}$k$ & Label\\
  \hline
  \hline
  \small$f(s) = s'$ & \\\small$ g(\uparrow, \downarrow, \leftarrow, \rightarrow)  = (\downarrow, \uparrow, \rightarrow, \leftarrow)$ & TRSAI\\\small$l(s') = s$  & \\
  \hline
  \small$f(s) = s$ & \\\small$ g(\uparrow, \downarrow, \leftarrow, \rightarrow)  = (\downarrow, \uparrow, \rightarrow, \leftarrow)$ & SDAI\\\small$l(s') = s'$  & \\
  \hline
  \small$f(s) = s$ & \\\small$ g(\uparrow, \downarrow, \leftarrow, \rightarrow)  = (\downarrow, \uparrow, \rightarrow, \leftarrow)$ & ODAI\\\small$l(s') = s' \mp (2, 0) \lor (0, 2)$  &\\
  \hline
  \small$f(s) = s$ & \\\small$ g(\uparrow, \downarrow, \leftarrow, \rightarrow)  = (\rightarrow, \leftarrow, \uparrow, \downarrow)$ & ODWA\\\small$l(s') = s' \mp (2, 0) \lor (0, 2)$  & \\
  \hline
  \small$f(s) = s'$ & \\\small$ g(\uparrow, \downarrow, \leftarrow, \rightarrow)  = (\uparrow, \downarrow, \leftarrow, \rightarrow)$ & TI\\\small$l(s') = s' \pm (1, 0) \lor (0, 1)$  & \\
  \hline
  \small$f(s) = s'$ & \\\small$ g(\uparrow, \downarrow, \leftarrow, \rightarrow)  = (\uparrow, \downarrow, \leftarrow, \rightarrow)$ & TIOD\\\small$l(s') = s$  & \\
  \hline
\end{tabularx}
\end{table}
\normalsize
The said transformations are resumed in the Table \ref{tab:grid-transf} along with the results in Table \ref{tab:results}.

With the scope of showing that augmenting the data set with the image of the symmetry $k$ leads to better model we compute the distributional shift between the true model $T$ and $\hat{T}$, the one learnt from $\mathcal{D}$, as the sum over every possible state-action pair of the Total Variation Distance for each transition.\\
In other words let
\begin{equation}
d(T, \hat{T}) = \dfrac{1}{2}\sum_{(s,s')\in S^2, a \in A}\lvert T(s,a,s') - \hat{T}(s,a,s')\rvert
\end{equation}
be the distance between $T$ and $\hat{T}$ and let
\begin{equation}
d(T, \hat{T}_k) = \dfrac{1}{2}\sum_{(s,s')\in S^2, a \in A}\lvert T(s,a,s') - \hat{T}_k(s,a,s')\rvert
\end{equation}
where $\hat{T}_k$ is the dynamics inferred from $\mathcal{A}_k$,
we report $\overline{\Delta}$ in Table \ref{tab:results}: the difference
\begin{equation}
\Delta = d(T, \hat{T}) - d(T, \hat{T}_k)
\end{equation}
averaged over $N=50$ simulations, showing that when $k$ is a true symmetry of the dynamics the said quantity is positive ($d(T, \hat{T}_k) < d(T, \hat{T})$) and when $k$ is not a real symmetry it is negative. Hence, augmenting the data set after having detected a true symmetry leads to better model learning, but the false detection of a symmetric transformation could also result in less representative models.
\vspace{-1mm}
\subsection{CartPole (Continuous)}
\vspace{-2mm}
As stated in the Introduction, the dynamics of CartPole is invariant with respect to the transformation $k=(f(s)= -s, g(a)=-a, l(s')=-s) \forall (s,s')\in S^2$.
In order to use Algorithm \ref{algo:cont} we first map the actions to real numbers: $\leftarrow = -1.5$ and $\rightarrow = 1.5$. We then normalize every state feature in the range $[-1.5, 1.5]$.
We tested Algorithm \ref{algo:cont} with four different alleged transformations $h$ over $N = 5$ different simulations, a batch of size $|\mathcal{D}|=10^3$ collected with a random policy, quantile order to compute the thresholds $q = 0.1$.
\begin{enumerate}
    \item \textbf{State and action reflection with respect to an axis in $x=0$ (SAR)}. Assuming that $\leftarrow$ is the reverse of $\rightarrow$ we proposed the following transformation: $k = \big(f(s) = -s$, $g(\leftarrow, \rightarrow) = (\rightarrow, \leftarrow)$, $l(s') =  -s'\big)$.
    The symmetric transformation is present in the batch with a percentage $\overline{\nu}_k = 0.80 \pm 0.05$.
    \item \textbf{Initial State Reflection (ISR).} We then tried the same transformation as before but without reflecting the next state $s'$: $k = \big(f(s) =  -s$, $g(\leftarrow, \rightarrow) = (\rightarrow, \leftarrow)$, $l(s') =  s'\big)$. This is clearly not a symmetry because $\overline{\nu}_k = 0.00 \pm 0.00$.
    \item \textbf{Action Inversion (AI).} What about reversing only the actions? $k = \big(f(s) =  s$, $g(\leftarrow, \rightarrow) = \paren{\rightarrow, \leftarrow}$, $l(s') =  s'\big)$ and $\overline{\nu}_k = 0.00 \pm 0.00$. Therefore this is not a symmetry of the environment according to our criterion.
    \item \textbf{Single Feature Inversion (SFI).} We also tried to reverse only one single feature of the starting state: $k = \big(f(x, v, \alpha, \omega) = (-x, v, \alpha, \omega)$, $g\paren{\leftarrow, \rightarrow} = \paren{\leftarrow, \rightarrow}$, $l(s') =  s'\big)$. In this case $\overline{\nu}_k = 0.05 \pm 0.03$.
    \item \textbf{Translation Invariance (TI).} We translated the position of the initial state $x$ and that of the final state $x'$ by an arbitrary value ($0.3$): $k = \big(f(x, v, \alpha, \omega) = (x+0.3, v, \alpha, \omega)$, $g(\leftarrow, \rightarrow) = \paren{\leftarrow, \rightarrow}$, $l(x', v', \alpha', \omega') = (x'+0.3, v', \alpha', \omega')\big)$. The percentage $\nu_{k} = 0.25 \pm 0.23$. Given the high variance this could be detected as a symmetry or not depending on the simulation.
\end{enumerate}
The proposed transformations are resumed in Table \ref{tab:cartpole-transf} along with the results in Table \ref{tab:results}.
\begin{table}[th]
\caption{CartPole. Proposed transformations and label.}\label{tab:cartpole-transf} \centering
\begin{tabularx}{\columnwidth}{c Y}
  \hline
  \noalign{\vskip 0.1mm}$k$ & Label\\
    \hline
    \hline
  \small$f(s) = -s$ &   \\\small$g(\leftarrow, \rightarrow) = (\rightarrow, \leftarrow)$ & SAR \\\small$l(s') = -s'$ &  \\
    \hline
  \small$f(s) = -s$ &  \\\small$g(\leftarrow, \rightarrow) = (\leftarrow, \rightarrow)$ & ISR \\\small$l(s') = s'$  & \\
    \hline
  \small$f(s) = s$ & \\\small$g(\leftarrow, \rightarrow) = (\rightarrow, \leftarrow)$ & AI \\\small$l(s') = s'$  & \\
    \hline
  \small$f((x, ...)) = (-x, ...)$ & \\\small$g(\leftarrow, \rightarrow) = (\leftarrow, \rightarrow)$ & SFI\\\small$l(s') = s'$  & \\
      \hline
    \small$f((x, ...)) = (x+0.3, ...)$ & \\\small$g(\leftarrow, \rightarrow) = (\leftarrow, \rightarrow)$ & TI\\\small$l((x', ...)) = (x'+0.3, ...)$  & \\
        \hline
\end{tabularx}
\end{table}

With the aim of computing the distributional shift we fit over $\mathcal{D}$ the evolution $\hat{s'}(s,a)$ with a Multi Layer Perceptron by minimizing the Mean Squared Error (MSE) between $\hat{s'}$ and ${s}$. We reiterate the same procedure over $\mathcal{D} \cup \mathcal{D}_k$ obtaining a next state
$\hat{s}'_k(s,a)$. We then randomly generate $10^6$ other transitions building up an evaluation batch $\mathcal{D}_{ev}$ from the environment and we evaluate with respect to the MSE these two models over the new data set.
We report in Table \ref{tab:results} the difference:
\begin{equation}
    \Delta = MSE_{\mathcal{D}_{ev}}(s', \hat{s}'(s,a)) - MSE_{\mathcal{D}_{ev}}(s', \hat{s}'_k(s,a))
\end{equation}
Again, when $\Delta > 0$ augmenting the data set leads to better models $\hat{s}'_k$, but when $\Delta < 0$ exploiting the transformation $h$ is detrimental to the dynamics.
\vspace{-1mm}
\subsection{Acrobot (Continuous)}
\begin{figure}
\centering
    \begin{tikzpicture}
    \draw[dashed] (0.05, -0.08) -- (-0.64, 0.65);
    \draw[dashed] (-1.4, 1.35) -- (-1.4, 0.7);
    \draw[rotate=45] (-0.05, 1) -- (-0.05, 2);
    \draw[rotate=45] (0.05, 1) -- (0.05, 2);
    \draw[rotate=45] (-0.05, 1) -- (-0.05, 2);
    \draw[line width=0.7mm] (-0.68, 0.65) -- (-0.4, -0.2) ;
    \draw[rotate=45] (0, 1) circle (0.05);
    \draw (-1.4, 1.4) circle (0.05);
    \draw[thick, dashed, ->] (-1.4, 0.8) arc (270:296:1) node[below=0.35cm, left=0.5cm]{$\alpha_1>0$};
    \draw[thick, dashed, <-] (-0.08, 0.1) arc (310:288:1) node[below right]{$\alpha_2<0$};
    \end{tikzpicture}
    \caption{Representation of a state of the Acrobot environment.}\label{fig:acrobot}
\end{figure}
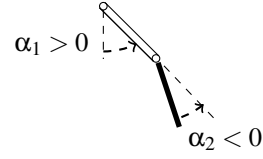

\vspace{-2mm}
The Acrobot environment consists of two poles linked with a rotating joint at one end. One of the poles is pinned to a wall with a second rotating joint (see Figure \ref{fig:acrobot}).
The system is affected by gravity and hence the poles are hanging down. An agent can apply a negative torque to the lower pole ($a = -1$), a positive one ($a=1$) or do nothing ($a = 0$). The goal is to push the lower pole as high as possible. The state consists of the sine and cosine of the two rotational joint angles ($\alpha_1, \alpha_2$) and the joint angular velocities ($\omega_1, \omega_2$) : $s = (\sin{\alpha_1}, \cos{\alpha_1}, \sin{\alpha_2}, \cos{\alpha_2}, \omega_1, \omega_2)$. The dynamics is invariant under the transformation $k = (f((\alpha_1, \alpha_2, \omega_1, \omega_2)) = (-\alpha_1, -\alpha_2, -\omega_1, -\omega_2)$ and $g(a) = -a, l((\alpha'_1, \alpha'_2, \omega'_1, \omega'_2)) = (-\alpha'_1, -\alpha'_2, -\omega'_1, -\omega'_2)) \forall (s,s') \in S^2$. In order to apply Algorithm \ref{algo:cont} we first normalize the state features and the action in the interval $[-3, 3]$. 
We tested Algorithm \ref{algo:cont} with four different alleged transformations $k$ over $N = 5$ different simulations, a batch of size $|\mathcal{D}|=10^3$ collected with a random policy, quantile order to compute the thresholds $q = 0.1$. The label of the transformations here after explained are resumed in Table \ref{tab:acrobot-transf}.
\begin{table}[t!]
\caption{Acrobot. Proposed transformations and label.}\label{tab:acrobot-transf} \centering
\begin{tabularx}{\columnwidth}{c Y}
  \hline
  \noalign{\vskip 0.1mm}$k$ & Label\\
    \hline
    \hline
  \small$f((s_1, s_2, \omega_1, \omega_2, \dots)) = -(s_1, s_2, \omega_1, \omega_2, \dots)$ & \\\small$g(-1, 0, 1) = (1, 0, -1)$ & AAVI \\\small$l((s'_1, s'_2, \omega'_1, \omega'_2, \dots)) = -(s'_1, s'_2, \omega'_1, \omega'_2, \dots)$  & \\
    \hline
    \small$f((c_1, c_2, \omega_1, \omega_2, \dots)) = -(c_1, c_2, \omega_1, \omega_2, \dots)$ & \\\small$g(-1, 0, 1) = (1, 0, -1)$ & CAVI \\\small$l((c'_1, c'_2, \omega'_1, \omega'_2, \dots)) = -(c'_1, c'_2, \omega'_1, \omega'_2, \dots)$  & \\
    \hline
    \small$f(s) = s$ & \\\small$g(-1, 0, 1) = (1, 0, -1)$ & AI\\\small$l(s') = s'$  & \\
    \hline
      \small$f(s) = -s$ & \\\small$g(-1, 0, 1) = (-1, 0, 1)$ & SSI\\\small$l(s') = s'$  & \\
        \hline

\end{tabularx}
\end{table}
\begin{table*}[!tbh]
\caption{Results of the algorithms in all environments. The columns on the right point to the tested alleged symmetry, indexed by its label as reported in Table \ref{tab:grid-transf} for the toroidal Grid, Table \ref{tab:cartpole-transf} for CartPole and Table \ref{tab:acrobot-transf} for Acrobot. All experiments were performed with $q = 0.1$. The true symmetries are displayed in bold.  }\label{tab:results}
\centering
\begin{tabularx}{\textwidth}{cc|ccccccc}
\cline{1-8}
\textit{\begin{tabular}[c]{ @{}c@{}} Environment\end{tabular}} &
  \textit{\begin{tabular}[c]{@{}c@{}} Metrics\end{tabular}} &
  \multicolumn{6}{c}{\textit{\begin{tabular}[c]{@{}c@{}} Alleged Transformation\end{tabular}}} &
   \\ \cline{1-8}
 &
   &
  \textbf{TRSAI} &
  SDAI &
  \textbf{ODAI} &
  ODWA &
  \textbf{TI} &
  TIOD &
   \\ \cline{3-8}
\multirow{2}{*}{Grid} &
  $\overline{\nu}_k$ &
  \scriptsize \textbf{0.6} $\pm$ \textbf{0.1} &
  \scriptsize $0.0 \pm 0.0$ &
  \scriptsize \textbf{0.4} $\pm$ \textbf{0.1} &
  \scriptsize $0.0 \pm 0.0$ &
  \scriptsize \textbf{0.5} $\pm$ \textbf{0.1} &
  \scriptsize $0.0 \pm 0.0$ &
   \\
 &
  $\overline{\Delta}$ &
  \scriptsize \textbf{27} $\pm$ \textbf{9} &
  \scriptsize -14 $\pm$ 3 &
  \scriptsize \textbf{47} $\pm$ \textbf{12} &
  \scriptsize -14 $\pm$ 3 &
  \scriptsize \textbf{39} $\pm$ \textbf{9} &
  \scriptsize -17 $\pm$ 2 &
   \\ \cline{1-8}
 &
   &
  \textbf{SAR} &
  ISR &
  AI &
  SFI &
  \textbf{TI} &
   &
   \\ \cline{3-7}
\multirow{2}{*}{CartPole} &
  $\overline{\nu}_k$ &
  \scriptsize \textbf{0.80} $\pm$ \textbf{0.05} &
  \scriptsize 0.00 $\pm$ 0.00 &
  \scriptsize 0.00 $\pm$ 0.00 &
  \scriptsize 0.05 $\pm$ 0.03 &
  \scriptsize \textbf{0.25} $\pm$ \textbf{0.23} &
   &
   \\
 &
  $\overline{\Delta}$ &
  \scriptsize \textbf{4} $\pm$ \textbf{2} $\bm{\times 10^{-4}}$ &
  \scriptsize -10 $\pm$ 1 $\times$ $10^{-2}$ &
  \scriptsize -8 $\pm$ 1 $\times$ $10^{-3}$ &
  \scriptsize -4 $\pm$ 3 $\times$ $10^{-3}$ &
  \scriptsize \textbf{3} $\pm$ \textbf{2} $\bm{\times10^{-4}}$ &
   &
   \\ \cline{1-8}
 &
   &
  \textbf{AAVI} &
  CAVI &
  AI &
  SSI &
   &
   &
   \\ \cline{3-6}
\multirow{2}{*}{Acrobot} &
  $\overline{\nu}_k$ &
  \scriptsize \textbf{0.86} $\pm$ \textbf{0.03} &
  \scriptsize $0.00 \pm 0.00$ &
  \scriptsize $0.35 \pm 0.09$ &
  \scriptsize $0.00 \pm 0.00$ &
   &
   &
   \\
 &
  $\overline{\Delta}$ &
  \scriptsize \textbf{3.3} $\pm$ \textbf{6.6} $\bm{\times 10^{-3}}$ &
  \scriptsize $-3.9 \pm 1.9 \times 10^{-2}$ &
  \scriptsize $-0.6 \pm 1.3 \times 10^{-2}$ &
  \scriptsize $-9.5 \pm 4.3 \times 10^{-2}$ &
   &
   &
   \\ \cline{1-8}
\end{tabularx}
\end{table*} 
\begin{enumerate}
    \item \textbf{Angles and Angular Velocities Inversion (AAVI).}\\
    $k = \big(f(\sin{\alpha_1}, \sin{\alpha_2},\cos{\alpha_1}, \cos{\alpha_2}, \omega_1, \omega_2) = (-\sin{\alpha_1}, -\sin{\alpha_2}, \cos{\alpha_1}, \cos{\alpha_2}, -\omega_1, -\omega_2),\\ g(a) = -a, l(\sin{\alpha'_1}, \sin{\alpha'_2}, \cos{\alpha'_1}, \cos{\alpha'_2}, \omega'_1, \omega'_2)\\ = (-sin{\alpha'_1}, -\sin{\alpha'_2}, \cos{\alpha'_1}, \cos{\alpha'_2}, -\omega'_1, -\omega'_2)\big)$. This transformation could be a symmetry since the percentage of images with a probability higher than the threshold is $\overline{\nu}_k = 0.86 \pm 0.03$.
    \item \textbf{Cosines and Angular Velocities Inversion (CAVI).}\\ 
    $k = \big(f(\sin{\alpha_1}, \sin{\alpha_2}, \cos{\alpha_1}, \cos{\alpha_2}, \omega_1, \omega_2) = (\sin{\alpha_1}, \sin{\alpha_2}, -\cos{\alpha_1}, -\cos{\alpha_2}, -\omega_1, -\omega_2),\\ 
    g(a) = -a, l(\sin{\alpha'_1}, \sin{\alpha'_2}, \cos{\alpha'_1}, \cos{\alpha'_2}, \omega'_1, \omega'_2)\\ = (\sin{\alpha'_1}, \sin{\alpha'_2}, -cos{\alpha'_1}, -\cos{\alpha'_2}, -\omega'_1, -\omega'_2)\big)$. This is not a symmetry since $\overline{\nu}_k = 0.00 \pm 0.00$.
    \item \textbf{Action Inversion (AI).} $k = \big(f(s) =  s, g(a) = -a, l(s') =  s'\big)$. In this case the results are not so indicative because $\overline{\nu}_k = 0.35 \pm 0.09$ and therefore the transformation could be falsely detected as a symmetry if a low threshold $\nu$ is chosen.
    \item \textbf{Starting State Inversion (SSI).} $k = \big(f(s) =  -s, g(a) = a, l(s') =  s'\big)$. This is clearly not a symmetric transformation because $\overline{\nu}_k = 0.00 \pm 0.00$.
\end{enumerate}


\vspace{-1mm}
The results can be found in Table \ref{tab:results} along with model performance. Using the said metrics $\Delta$ to evaluate the gains of data augmentation, we notice that in this case the high dimensionality of the state space makes it harder for the MLP to reconstruct the next state and that even the discovery of the true symmetry $k = (f((\alpha_1, \alpha_2, \omega_1, \omega_2)) = -(\alpha_1, \alpha_2, \omega_1, \omega_2), g(a) = -a, l((\alpha'_1, \alpha'_2, \omega'_1, \omega'_2)) = -(\alpha'_1, \alpha'_2, \omega'_1, \omega'_2))$ seems to not lead to the expected results because the confidence interval of $\overline{\Delta}$ also falls on the negative part of the real axis.

\vspace{-3mm}
\subsection{Discussion}
\vspace{-3mm}
In the light of the results we can state that, when used with care, the proposed approaches manage to detect the alleged transformations of the transitions which are symmetries of the dynamics.
However, note that the present experiments in the continuous environments were performed with a quantile of order $q=0.1$. The outcomes can greatly be affected by the choice of $q$ and further studies should be carried out in the next future to address this limitation.
Moreover, the expert choice of $\nu$ can also have a huge impact on the final detection. Indeed, $\nu=0.30$ in the Acrobot environment could lead to the misdetection of AI as a symmetry, while the TI symmetry of CartPole requires a very low ($\nu<0.20$) in order to be detected (Table \ref{tab:results}).
\vspace{-6mm}
\section{\uppercase{Conclusions}}
\vspace{-3mm}
\label{sec:conclusion}
The present work proposed an expert-guided approach to data augment the batch of trajectories used to learn a model in Offline Model Based RL. More specifically, new information is created assuming that the dynamics of the system is invariant with respect to some symmetric transformation of the state and action representations. The alleged symmetry is detected exploiting both expert knowledge and the fulfillment of an a priori established criterion. Experimental results in proof-of-concepts deterministic environments suggest that the proposed method could be an useful tool for the developer of Reinforcement Learning algorithms when he is ought to apply them to real world scenarios. Nevertheless, the strong dependency from preprocessing, from neural network hyperparameters to be fine tuned before training and the from the choice of the quantile order $q$ and of the percentage threshold $\nu$ impose that Algorithms \ref{algo:disc} and \ref{algo:cont} should be handled with care. Indeed, the false detection of a symmetry and the following data augmenting could lead to catastrophic aftermaths when the retrieved optimal policy is going to be applied to the real environment.

Future research perspectives involve (1) using more recent Normalizing Flows approaches to perform density estimation, e.g. we do not exclude that state-of-the art continuous time Normalizing Flows who exploit ODE solvers like FFJORD \cite{DBLP:conf/iclr/GrathwohlCBSD19} might yield even better results; (2) extending the approach to also tackle stochastic environments.

\vspace{-4mm}
\section*{ACKNOWLEDGEMENTS}
\vspace{-2mm}
This work was carried out during a visit at the Pompeu Fabra University (UPF) in Barcelona and was funded by the Artificial and Natural Intelligence Toulouse Institute (ANITI) - Institut 3iA (ANR-19-PI3A-0004). We thank Hector Geffner, Anders Jonsson and the Artificial Intelligence and Machine Learning group of the UPF for their warm hospitality.
\newpage
\bibliographystyle{apalike}
{\small
\bibliography{references}}

\begin{thebibliography}{}

\bibitem[Abel et~al., 2020]{abel2020value}
Abel, D., Umbanhowar, N., Khetarpal, K., Arumugam, D., Precup, D., and Littman,
  M. (2020).
\newblock Value {Preserving} {State-Action} {Abstractions}.
\newblock In {\em International Conference on Artificial Intelligence and
  Statistics}, pages 1639--1650. PMLR.

\bibitem[Bellman, 1966]{bellman1966dynamic}
Bellman, R. (1966).
\newblock Dynamic {Programming}.
\newblock {\em Science}, 153(3731):34--37.

\bibitem[Brockman et~al., 2016]{brockman2016openai}
Brockman, G., Cheung, V., Pettersson, L., Schneider, J., Schulman, J., Tang,
  J., and Zaremba, W. (2016).
\newblock {OpenAI} {Gym}.
\newblock {\em arXiv preprint arXiv:1606.01540}.

\bibitem[Castro, 2020]{castro2020scalable}
Castro, P.~S. (2020).
\newblock Scalable {Methods} for {Computing} {State} {Similarity} in
  {Deterministic} {Markov} {Decision} {Processes}.
\newblock In {\em Proceedings of the AAAI Conference on Artificial
  Intelligence}, volume~34, pages 10069--10076.

\bibitem[Dean and Givan, 1997]{dean1997model}
Dean, T. and Givan, R. (1997).
\newblock Model {Minimization} in {Markov} {Decision} {Processes}.
\newblock In {\em AAAI/IAAI}, pages 106--111.

\bibitem[Dinh et~al., 2015]{DBLP:journals/corr/DinhKB14}
Dinh, L., Krueger, D., and Bengio, Y. (2015).
\newblock {NICE:} {Non-linear} {Independent} {Components} {Estimation}.
\newblock In Bengio, Y. and LeCun, Y., editors, {\em 3rd International
  Conference on Learning Representations, {ICLR} 2015, San Diego, CA, USA, May
  7-9, 2015, Workshop Track Proceedings}.

\bibitem[Ferns et~al., 2004]{ferns2004metrics}
Ferns, N., Panangaden, P., and Precup, D. (2004).
\newblock Metrics for {Finite} {Markov} {Decision} {Processes}.
\newblock In {\em UAI}, volume~4, pages 162--169.

\bibitem[Givan et~al., 2003]{givan2003equivalence}
Givan, R., Dean, T., and Greig, M. (2003).
\newblock Equivalence {Notions} and {Model} {Minimization} in {Markov}
  {Decision} {Processes}.
\newblock {\em Artificial Intelligence}, 147(1-2):163--223.

\bibitem[Grathwohl et~al., 2019]{DBLP:conf/iclr/GrathwohlCBSD19}
Grathwohl, W., Chen, R. T.~Q., Bettencourt, J., Sutskever, I., and Duvenaud, D.
  (2019).
\newblock {FFJORD:} {Free-Form} {Continuous} {Dynamics} for {Scalable}
  {Reversible} {Generative} {Models}.
\newblock In {\em 7th International Conference on Learning Representations,
  {ICLR} 2019, New Orleans, LA, USA, May 6-9, 2019}. OpenReview.net.

\bibitem[Gross, 1996]{Gross14256}
Gross, D.~J. (1996).
\newblock The role of symmetry in fundamental physics.
\newblock {\em Proceedings of the National Academy of Sciences},
  93(25):14256--14259.

\bibitem[Kobyzev et~al., 2020]{kobyzev2020normalizing}
Kobyzev, I., Prince, S., and Brubaker, M. (2020).
\newblock Normalizing {Flows}: {An} {Introduction} and {Review} of {Current}
  {Methods}.
\newblock {\em IEEE Transactions on Pattern Analysis and Machine Intelligence}.

\bibitem[Levine et~al., 2020]{Levine2020OfflineRL}
Levine, S., Kumar, A., Tucker, G., and Fu, J. (2020).
\newblock Offline reinforcement learning: Tutorial, review, and perspectives on
  open problems.
\newblock {\em ArXiv}, abs/2005.01643.

\bibitem[Li et~al., 2006]{li2006towards}
Li, L., Walsh, T.~J., and Littman, M.~L. (2006).
\newblock Towards a {Unified} {Theory} of {State} {Abstraction} for {MDPs}.
\newblock {\em ISAIM}, 4:5.

\bibitem[Mandel et~al., 2016]{mandel2016efficient}
Mandel, T., Liu, Y.-E., Brunskill, E., and Popovic, Z. (2016).
\newblock Efficient {Bayesian} {Clustering} for {Reinforcement} {Learning}.
\newblock In {\em IJCAI}, pages 1830--1838.

\bibitem[Mausam and Kolobov, 2012]{kolobov2012planning}
Mausam and Kolobov, A. (2012).
\newblock {\em Planning with Markov Decision Processes: An AI Perspective}.
\newblock Morgan \& Claypool Publishers.

\bibitem[Narayanamurthy and Ravindran, 2008]{narayanamurthy2008hardness}
Narayanamurthy, S.~M. and Ravindran, B. (2008).
\newblock On the {Hardness} of {Finding} {Symmetries} in {Markov} {Decision}
  {Processes}.
\newblock In {\em Proceedings of the 25th international conference on Machine
  learning}, pages 688--695.

\bibitem[Ravindran and Barto, 2001]{ravindran2001symmetries}
Ravindran, B. and Barto, A.~G. (2001).
\newblock Symmetries and {Model} {Minimization} in {Markov} {Decision}
  {Processes}.
\newblock Technical report, USA.

\bibitem[Ravindran and Barto, 2004]{ravindran2004approximate}
Ravindran, B. and Barto, A.~G. (2004).
\newblock Approximate {Homomorphisms}: {A} {Framework} for {Non-exact}
  {Minimization} in {Markov} {Decision} {Processes}.

\bibitem[Ruan et~al., 2015]{ruan2015representation}
Ruan, S.~S., Comanici, G., Panangaden, P., and Precup, D. (2015).
\newblock Representation {Discovery} for {MDPs} {Using} {Bisimulation}
  {Metrics}.
\newblock In {\em Twenty-Ninth AAAI Conference on Artificial Intelligence}.

\bibitem[Sutton, 1990]{SUTTON1990216}
Sutton, R.~S. (1990).
\newblock Integrated {A}rchitectures for {L}earning, {P}lanning, and {R}eacting
  {B}ased on {A}pproximating {D}ynamic {P}rogramming.
\newblock In Porter, B. and Mooney, R., editors, {\em Machine Learning
  Proceedings 1990}, pages 216--224. Morgan Kaufmann, San Francisco (CA).

\bibitem[Taylor et~al., 2009]{NIPS2008_6602294b}
Taylor, J., Precup, D., and Panagaden, P. (2009).
\newblock Bounding {Performance} {Loss} in {Approximate} {MDP} {Homomorphisms}.
\newblock In Koller, D., Schuurmans, D., Bengio, Y., and Bottou, L., editors,
  {\em Advances in Neural Information Processing Systems}, volume~21. Curran
  Associates, Inc.

\bibitem[van~der Pol et~al., 2020a]{10.5555/3398761.3398926}
van~der Pol, E., Kipf, T., Oliehoek, F.~A., and Welling, M. (2020a).
\newblock Plannable {Approximations} to {MDP} {Homomorphisms}: {Equivariance}
  under {Actions}.
\newblock In {\em Proceedings of the 19th International Conference on
  Autonomous Agents and MultiAgent Systems}, AAMAS '20, page 1431–1439,
  Richland, SC. International Foundation for Autonomous Agents and Multiagent
  Systems.

\bibitem[van~der Pol et~al., 2020b]{NEURIPS2020_2be5f9c2}
van~der Pol, E., Worrall, D., van Hoof, H., Oliehoek, F., and Welling, M.
  (2020b).
\newblock {MDP} {Homomorphic} {Networks}: {Group} {Symmetries} in
  {Reinforcement} {Learning}.
\newblock In Larochelle, H., Ranzato, M., Hadsell, R., Balcan, M.~F., and Lin,
  H., editors, {\em Advances in Neural Information Processing Systems},
  volume~33, pages 4199--4210. Curran Associates, Inc.

\end{thebibliography}
\end{document}